\documentclass[fleqn,10pt]{wlscirep}
\usepackage[utf8]{inputenc}
\usepackage[T1]{fontenc}
\usepackage{lineno}
\usepackage{times}
\usepackage{epsfig}
\usepackage{graphicx}
\usepackage{caption}
\usepackage{subcaption}
\usepackage{amsmath}
\usepackage{amssymb}
\usepackage{mathtools}
\usepackage{caption}
\usepackage{subcaption}
\usepackage{xr}

\usepackage{lineno}
\usepackage{lscape}
\usepackage{booktabs}
\usepackage{amsmath}
\usepackage{xr}
\usepackage{rotating}
\usepackage{multirow}
\usepackage{graphics}
\usepackage{rotating}
\usepackage{verbatim}
\usepackage{float} % [H]

\newcommand{\subD}[1]{$_{\text{#1D}}$}

\title{When Segmentation is Not Enough: Rectifying Visual-Volume Discordance Through Multisensor Depth-Refined Semantic Segmentation for Food Intake Tracking in Long-Term Care}

\author[1,2,3*+]{Kaylen~J.~Pfisterer}
\author[4+]{Robert~Amelard}
\author[1,2]{Audrey~G.~Chung}
\author[5]{Braeden~Syrnyk}
\author[1,2]{Alexander~MacLean}
\author[3,6]{Heather~H.~Keller}
\author[1,2,3]{Alexander~Wong}
\affil[1]{University of Waterloo, Waterloo, Systems Design Engineering,  Waterloo, ON, N2L 3G1, Canada}
\affil[2]{Waterloo AI Institute, Waterloo, ON, N2L 3G1, Canada}
\affil[3]{Schlegel-UW Research Institute for Aging, Waterloo, N2J 0E2, Canada}
\affil[4]{KITE-Toronto Rehabilitation Institute, University Health Network, Toronto, ON M5G 2A2}
\affil[5]{University of Waterloo, Waterloo, Mechanical and Mechatronics Engineering,  Waterloo, ON, N2L 3G1, Canada}
\affil[6]{University of Waterloo, Waterloo, Kinesiology, Waterloo, ON, N2L 3G1, Canada}

\affil[*]{kpfisterer@uwaterloo.ca}

\affil[+]{these authors contributed equally to this work}


\begin{abstract}
Malnutrition is a multidomain problem affecting 54\% of older adults in long-term care (LTC). Monitoring nutritional intake in LTC is laborious and subjective, limiting clinical inference capabilities. Recent advances in automatic image-based food estimation have not yet been evaluated in LTC settings. Here, we describe a fully automatic imaging system for quantifying food intake. We propose a novel deep convolutional encoder-decoder food network with depth-refinement (EDFN-D) using an RGB-D camera for quantifying a plate’s remaining food volume relative to reference portions in whole and modified texture foods. We trained and validated the network on the pre-labelled UNIMIB2016 food dataset and tested on our two novel LTC-inspired plate datasets (689 plate images, 36 unique foods). EDFN-D performed comparably to depth-refined graph cut on IOU (0.879 vs. 0.887), with intake errors well below typical 50\% (mean percent intake error: -4.2\%). We identify how standard segmentation metrics are insufficient due to visual-volume discordance, and include volume disparity analysis to facilitate system trust. This system provides improved transparency, approximates human assessors with enhanced objectivity, accuracy, and precision while avoiding hefty semi-automatic method time requirements. This may help address short-comings currently limiting utility of automated early malnutrition detection in resource-constrained LTC and hospital settings.
\end{abstract}
\begin{document}

\flushbottom
\maketitle

\thispagestyle{empty}

\section*{Introduction}

Malnutrition has multidomain effects and should be monitored especially in older adults as evidenced by the clinical ramifications (morbidity~\cite{pirlich2001}, decreased quality of life~\cite{keller2004}), annual economic impact (USA \$15.5 billion~\cite{goates2016}, UK \pounds 7.3 billion~\cite{russell2007}), and high prevalence (23\% malnourished~\cite{kaiser2010}). In long-term care (LTC) homes, the prevalence is greater with malnutrition or risk for malnutrition affecting 54\% of residents~\cite{keller2019prevalence}. Malnutrition and risk for malnutrition is primarily due to low food intake of residents~\cite{keller2017prevalence}. Thus, tracking and preventing poor food intake is paramount. However, we lack quality tracking methods for food and fluid intake, especially needed with multiple staff involved in the care of residents over the course of a day or week. While methods for measuring food and fluid intake exist, these methods are subject to self-reporting bias, negatively affecting both validity and accuracy~\cite{martin2008,williamson2003} (e.g., errors up to: 400\%, 24-hr recall; 50\%, portion size~\cite{bingham1991}). In the LTC sector, personnel are mandated to report at-risk resident's food and fluid intake; however, correct estimation of intake occurs only 44\% of the time under routine conditions and as low as 38\% of the time with delayed recording~\cite{castellanos2002}. As a result, trust in these measurements is low, with limited utility in practice but care providers would like to utilize this information if measurement reliability and trust in these measurements could be ensured~\cite{pfisterer2019}. 

Automated tools may provide a time efficient and cost effective, and objective alternative. We previously established that the LTC sector requires a system that is reliable, accurate, cost effective and time efficient for measuring food and resulting energy, macro and micronutrient intake~\cite{pfisterer2019}. However, before assessing food and fluid intake, three main questions must first be answered: \textit{where} is there food (segmentation), \textit{what} foods are present (classification), and \textit{how much} food remains relative to the initial amount (volume estimation)? For the purpose of this paper, we focus on the \textit{where} (segmentation), and the \textit{how much} (volume estimation), as types of food items in LTC are well constrained through monthly menu-planning. Additionally, food classification has received the most attention; reviews of the literature can be found elsewhere (e.g.,~\cite{lo2020image,bruno2017survey,doulah2019systematic,pouladzadeh2016,subhi2019}).

Food segmentation is a domain which is relatively unexplored. Existing food intake tracking systems rely on images from multiple perspectives~\cite{kong2015dietcam,pouladzadeh2014measuring}, require a single image with a fiducial marker (i.e., reference  object~\cite{okamoto2016,zhu2015}), or may not be suitable for real-time monitoring(~\cite{meyers2015}). Others are limited to predicting food areas with a bounding box~\cite{shimoda2015cnn}, require manual labelling for each food item~\cite{meyers2015}, or require manual selection of bounding boxes~\cite{kawano2015}. These methods involve operator time and may impact accuracy. For example, two operators  may segment food differently, foods may be incorrectly labelled, or labels may be missed in some cases. One semi-automatic method, interactive graph cut segmentation, has yielded strong accuracy in food segmentation~\cite{pouladzadeh2016food,hassannejad2017,kawano2015}. It does not impart the same degree of burden as manual segmentation and we consider this as an ``applied  ground truth''. However, interactive annotation graph cut~\cite{boykov2001graphcut} requires user input to initialize the segmentation process (e.g., drawing areas to keep or discard). Adding a few seconds per image within the LTC environment makes it prohibitive within this context. 

While food image segmentation progress has been made, error assessment in these systems tends not to be reported, or segmentation is coupled with either classification~\cite{pouladzadeh2016food,kawano2015,he2013food,meyers2015,shimoda2015cnn,wang2018,yunus2018framework} or volume~\cite{hassannejad2017}, making sources of error difficult to disentangle. This has practical implications as there is generally no way to systematically assess error propagation as part of the pipeline for predicting nutritional outcomes. This results in the system operating as a ``black-box'', which may limit the uptake of these approaches in practice due to low perceived trust-worthiness. Beyond the user and ethical perspectives, several researchers also describe the need for accurate segmentation methods for accurately predicting nutritional information (e.g., energy, macro-/micronutrient content) down-stream in the pipeline~\cite{meyers2015,aslan2018semantic,wang2018,zhu2015}. Intersection over union (IOU) has been the most consistently reported accuracy metric~\cite{zheng2018image,ciocca2019evaluating,aslan2018semantic,aguilar2018grab}. IOU has several advantages over more traditional precision/recall metrics as it considers the proportion of properly assigned pixels but also penalizes false positive predictions; however, it may not necessarily capture an assessor's perspective on what is relevant food (e.g., to include or not include crumbs).

Food volume estimation systems are also relevant for estimating portion sizes of consumed food by subtracting the remainder of food from the original portion size. Several attempts have been made using template shape matching~\cite{he2013food,xu2013,chae2011,jia2014accuracy,ofei2019validation,rachakonda2020ilog,herzig2020volumetric}. Three main drawbacks of these methods are the requirement of a shape library, difficulty with template matching with occlusions, and varying preparation methods of the same food. For example, if a food is prepared differently it may not map onto the appropriate shape model (i.e., 3D banana in peel may be in the library but sliced or diced banana may not). Similar to the segmentation problem, others have applied a multi-image perspective or stereo reconstruction for volume estimation~\cite{dehais2017,puri2009recognition} or building a 3D representation through point-cloud representation~\cite{rahman2012food}. The main drawback of these approaches is the time required to take the photos from different perspectives or gather enough sample points scanned for an accurate 3D representation; lack of time is a main concern when considering the LTC context. Another challenge of these approaches is accurate modelling/measuring of highly textured foods. This is a particularly salient issue for LTC where modified textures are often prescribed as part of a therapeutic diet so very different foods can appear similar (e.g., minced or pureed foods). Others have employed depth cameras and structured lighting to map the topology of the foods~\cite{fang2016,shang2011,meyers2015,chen2012,liao2016}. One drawback of depth- and structured-light-only methods is highly reflective foods (e.g., gelatin, soup) which can throw off the readings but they have an advantage at being more robust against illumination variations~\cite{liao2016}. Despite the deep learning boom, acquiring adequately large and complete food datasets for training and testing has limited progress~\cite{zhou2019application}. The few forays have been fairly early-stage as they contain either large mean volume estimation errors (up to 400~mL error~\cite{meyers2015}), measure in terms of niche units (bread units tailored to diabetes~\cite{ferdinand2017diabetes60}), or are limited to a small number of or synthetic food items in a highly controlled environment (e.g.,~\cite{lo2018food,lo2019point2volume}).

Recent reviews corroborate that further work is needed for developing an accurate, objective and cost-effective automated system~\cite{subhi2019,doulah2019systematic} and describe the need for more complex meal scenarios (i.e., beyond solid, separated foods, or synthetic foods)~\cite{pouladzadeh2016,subhi2019} and adequate statistical analyses of methods~\cite{doulah2019systematic}. In line with these opportunity areas and current human computer interaction trajectories~\cite{abdul2018trends}, we seek to improve trust and transparency by focusing on developing an explainable system which approximates human assessors but with enhanced objectivity, accuracy, and precision. Specifically in this paper, we describe and evaluate a novel fully automated depth-refined multisensor food intake tracking system. Here the depth-refined segmentation and volume estimation have been decoupled to disentangle and assess potential sources of error. Through this decoupling, we seek to enhance reliability for eventual integration with nutritional intake estimation, and to reduce potential barrier to uptake in practice. We designed the segmentation system to be used in clinical settings (such as LTC or hospitals) with acquisition consistent with LTC food and fluid intake visual assessment procedures. Our system is comprised of an RGB-D camera, a novel deep convolutional neural network encoder-decoder food network (EDFN), fuses output from the EDFN with superpixel processing, and incorporates depth information for enhanced segmentation and volume estimation.  The use of a single RGB-D camera brings simplicity over a multi-camera or multi-perspective set-up, reducing processing and acquisition time while removing subjectivity in the assessment. We trained our EDFN on the UNIMIB2016 dataset~\cite{cioccaJBHI} and test it on two novel datasets to reduce bias and enhance generalizability. The two novel datasets are 1) a regular texture foods dataset and, 2) a modified texture foods dataset (e.g., pureed, minced). To the best of our knowledge, this is the first modified texture foods dataset used for segmentation or volume estimation. We conducted comprehensive analyses including IOU, 2D- and 3D- percent intake error, absolute intake error, and mean intake error bias. We use ground-truth hand segmentation and comparison against an ``applied ground truth'' through the graph cut semi-automated method. We supplement these analyses with volume disparities to illuminate how segmentation strategies impact accuracy and under what conditions IOU may be insufficient to assess true accuracy. Using this more holistic construct for assessment, we aim to enable trust in the system and document potential circumstances and limitations of the system relevant to the LTC domain for early malnutrition detection via plate-by-plate food consumption tracking.

\section*{Results}

\label{results}

\begin{sidewaysfigure*}
    \centering
	\includegraphics[width=\textwidth]{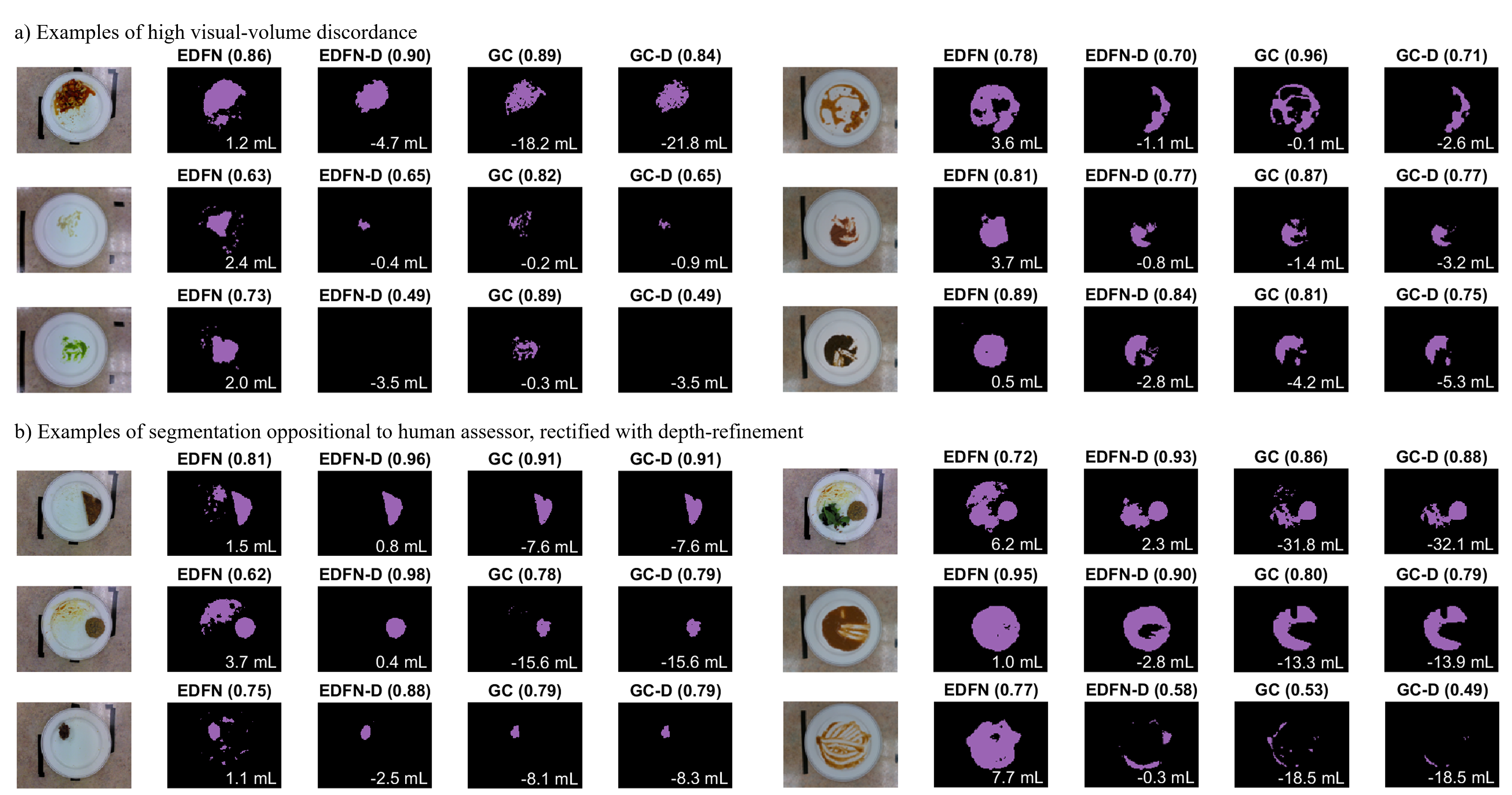}
    \caption{Visual comparison of proposed method (EDFN) and our ``applied ground truth'' (GC) both without and with depth-refinement (EDFN-D, GC-D). Examples span both regular and modified textures and illustrate a) plates with high visual-volume discordance, as well as b) examples where depth-refinement rectified oppositional segmentation compared to human assessor. IOU is shown in black bold at the top of the frame while volume error (mL) is indicated in white text at the bottom of each frame; negative volume error implies under-segmentation (i.e., over-estimation of intake).}
    \label{fig:vis_comp}
\end{sidewaysfigure*}

When solely considering intake using segmentation, the form factor of food inherently assumes depth is uniform across the segmented portion. However, this approach fails to fully capture the context of food; we refer to this as the ``\textbf{visual-volume discordance}''. For example, consider one tablespoon (15 mL) of tomato sauce, this sauce could be piled relatively high into a mound (i.e., representing relatively few plate pixels), or could be very thinly spread across the majority of the plate (i.e., representing many plate pixels). From a computer vision approach to segmentation, these two plates would yield extremely different segmented areas while the absolute volume of the sauce would be the same. A human assessor is able to note there is little sauce on the plate in either configuration. Depth-refinement, either as part of the segmentation pipeline or conducted through relative changes in volume for food volume intake assessment, circumvents this issue by providing context beyond the pixel count of a segment and brings assessment closer (but with higher precision and accuracy) to a human assessor. Now consider what is deemed ``ground truth'' from hand segmentation of an image. Here, human assessors indicate \textit{where} on the plate there was food to generate the ground truth. However, considering segmentation accuracy in isolation misses important context about \textit{how much} food is present, which is the more pertinent question for assessing food intake. As such, we cannot rely fully on classical segmentation accuracy for evaluating system performance since there can be a strong discordance between visual (RGB) and volume (RGB-D) assessments. Volume consideration is particularly essential for estimating food intake when accounting for the high prevalence of modified texture foods in LTC. While metrics pertaining to visual accuracy are most synonymous with traditional assessments of segmentation accuracy (e.g., IOU), metrics pertaining to volume accuracy may be more representative of true intake, especially for modified texture foods which have higher fluidity. Figure~\ref{fig:vis_comp} provides a visual analyses of the results taking into account these considerations while Table~\ref{tab:seg_vol_comp_analysis} provides a numerical summary of results. Data are reported as (mean $\pm$ SD).

\begin{sidewaystable}[]

\caption{Comparative analyses of system performance within and across LTC datasets between our proposed method (EDFN, EDFN-D) and the ``applied ground truth'' graph cuts (GC, GC-D). \% error intake refers to the proportion of segmented pixels calculated using the predicted estimate minus the target ground-truth hand segmented regions; 2D: no-depth, 3D: with depth-refinement.}

\label{tab:seg_vol_comp_analysis}
\begin{tabular}{l|ccc|cc|ccc}
\textit{\textbf{Dataset}} & \multicolumn{3}{c}{\textbf{Segmentation accuracy}} & \multicolumn{2}{c}{\textbf{Intake accuracy}} & \multicolumn{3}{c}{\textbf{Volume estimation accuracy (mL)}} \\
\textit{Method} & GSA & FSA & IOU & 2D \% intake error & 3D \% intake error & Mean absolute error & Mean error bias & Volume intake error \\ \hline
\multicolumn{3}{l}{\textit{\textbf{Regular texture foods}}} &  &  &  &  &  &  \\
\textit{EDFN} & 0.973 (0.018) & 0.943 (0.047) & 0.885 (0.069) & -23.4 (21.2) & -9.1 (8.8) & 17.1 (49.2) & -14.7 (50.0) & -129.2 (154.3) \\
\textit{EDFN-D} & 0.984 (0.012) & 0.918 (0.048) & 0.927 (0.029) & -16.5 (17.6) & -9.0 (8.9) & 18.0 (50.0) & -17.2 (50.3) & -130.2 (154.8) \\
\textit{GC} & 0.987 (0.009) & 0.955 (0.055) & 0.938 (0.036) & -6.8 (13.5) & 0.4 (1.3) & 4.5 (5.5) & -0.0 (7.1) & 1.8 (6.6) \\
\textit{GC-D} & 0.988 (0.005) & 0.933 (0.060) & 0.941 (0.029) & -5.3 (12.7) & 0.4 (1.3) & 4.6 (5.4) & -1.8 (6.9) & 0.2 (6.5) \\
\multicolumn{3}{l}{\textit{\textbf{Modified texture foods}}} &  &  &  &  &  &  \\
\textit{EDFN} & 0.990 (0.011) & 0.922 (0.165) & 0.846 (0.114) & -44.0 (38.6) & -0.8 (5.2) & 2.8 (3.1) & 1.7 (3.8) & 0.3 (3.6) \\
\textit{EDFN-D} & 0.991 (0.014) & 0.697 (0.348) & 0.819 (0.166) & -16.5 (20.0) & 1.4 (5.9) & 2.3 (3.2) & -0.7 (3.9) & 0.8 (3.6) \\
\textit{GC} & 0.995 (0.006) & 0.834 (0.157) & 0.898 (0.082) & -27.4 (25.5) & 0.7 (3.1) & 2.2 (2.7) & -1.9 (2.9) & -0.9 (3.3) \\
\textit{GC-D} & 0.991 (0.013) & 0.656 (0.328) & 0.819 (0.165) & -14.5 (18.1) & 2.1 (4.6) & 3.2 (3.4) & -3.1 (3.5) & -0.5 (3.4) \\
\textit{\textbf{All foods}} &  &  &  &  &  &  &  &  \\
\textit{EDFN} & 0.981 (0.017) & 0.934 (0.116) & 0.867 (0.094) & -32.8 (32.0) & -5.3 (8.5) & 10.8 (37.4) & -7.4 (38.2) & -71.4 (131.7) \\
\textit{EDFN-D} & 0.987 (0.013) & 0.819 (0.259) & 0.879 (0.125) & -16.5 (18.7) & -4.2 (9.2) & 11.0 (38.1) & -9.9 (38.4) & -71.8 (132.3) \\
\textit{GC} & 0.991 (0.008) & 0.901 (0.128) & 0.920 (0.064) & -16.2 (22.3) & 0.5 (2.3) & 3.5 (4.6) & -0.8 (5.7) & 0.6 (5.6) \\
\textit{GC-D} & 0.990 (0.010) & 0.809 (0.262) & 0.887 (0.127) & -9.5 (16.1) & 1.1 (3.3) & 4.0 (4.7) & -2.4 (5.7) & -0.1 (5.3) \\ \hline
\multicolumn{8}{l}{\begin{minipage}[t]{0.8\textwidth}Values are (mean $\pm$ SD) GSA: Global segmentation accuracy, FSA: Food segmentation accuracy, IOU: intersection over union.\end{minipage}}
\end{tabular}

\bigskip\bigskip
\caption{Summary of volume estimation accuracy across portion sizes for the modified texture foods dataset across our proposed EDFN and EDFN-D.}
\label{tab:portion_size_analysis}
\begin{tabular}{l|cc|cc|ccc}
\textit{\textbf{Dataset}} & \multicolumn{2}{c}{\textbf{Segmentation accuracy}} & \multicolumn{2}{c}{\textbf{Intake accuracy}} & \multicolumn{3}{c}{\textbf{Volume estimation accuracy (mL)}} \\
\textit{Portion} & GSA & FSA & 2D \% intake error & 3D \% intake error & Mean absolute error & Mean error & Volume intake error \\ \hline
\multicolumn{3}{l}{\textit{\textbf{EDFN   (no-depth-refinement)}}} &  &  &  &  &  \\
\textit{P1} & 0.996 (0.003) & 0.970 (0.066) & 0.0 (0.0) & 0.0 (0.0) & 3.3 (3.4) & 2.0 (4.3) & 0.0 (0.0) \\
\textit{P2} & 0.994 (0.006) & 0.965 (0.089) & -30.2 (17.2) & -0.8 (8.0) & 3.0 (3.1) & 2.1 (3.8) & -0.1 (3.3) \\
\textit{P3} & 0.992 (0.009) & 0.941 (0.125) & -50.1 (23.5) & -0.9 (5.3) & 2.9 (2.9) & 1.9 (3.6) & 0.1 (3.7) \\
\textit{P4} & 0.987 (0.013) & 0.912 (0.176) & -68.0 (35.3) & -1.1 (5.6) & 3.0 (3.2) & 1.4 (4.2) & 0.6 (4.5) \\
\textit{P5} & 0.982 (0.014) & 0.821 (0.254) & -71.9 (43.4) & -1.0 (3.8) & 2.0 (2.6) & 1.1 (3.1) & 0.9 (4.6) \\
\multicolumn{3}{l}{\textit{\textbf{EDFN-D   (depth-refined)}}} &  &  &  &  &  \\
\textit{P1} & 0.996 (0.004) & 0.945 (0.074) & 0.0 (0.0) & 0.0 (0.0) & 2.6 (3.3) & 0.1 (4.3) & 0.0 (0.0) \\
\textit{P2} & 0.994 (0.010) & 0.908 (0.132) & -22.8 (17.5) & 0.3 (8.7) & 1.9 (3.0) & 0.0 (3.6) & 0.1 (3.3) \\
\textit{P3} & 0.992 (0.013) & 0.824 (0.162) & -30.6 (17.5) & 1.4 (6.3) & 2.3 (3.5) & -0.5 (4.1) & 0.7 (4.2) \\
\textit{P4} & 0.987 (0.018) & 0.640 (0.275) & -25.2 (20.1) & 2.8 (6.6) & 2.9 (3.8) & -1.7 (4.5) & 1.8 (4.4) \\
\textit{P5} & 0.986 (0.016) & 0.157 (0.262) & -3.9 (16.2) & 2.3 (3.5) & 1.7 (2.2) & -1.5 (2.4) & 1.6 (4.0) \\ \hline
\multicolumn{8}{l}{\begin{minipage}[t]{0.72\textwidth}Values are (mean $\pm$ SD) GSA: Global segmentation accuracy, FSA: Food segmentation accuracy, IOU: intersection over union.\end{minipage}}
\end{tabular}
\end{sidewaystable}

\subsection*{Regular Texture Foods}
\label{sec:wholefood}
\subsubsection*{Food Segmentation Accuracy}
Regarding the regular texture foods dataset, food segmentation accuracy was high for our proposed system without and with depth-refinement (EDFN: 0.943~$\pm$~0.047, EDFN-D: 0.918~$\pm$~0.048). Our proposed, depth-refined system was comparable to the no-depth and depth-refined graph cut implementations (GC: 0.955~$\pm$~0.055, GC-D: 0.933$~\pm$ 0.060). Variance was similar but slightly lower for our proposed EDFN and EDFN-D than GC and GC-D.

\subsubsection*{Segmentation Agreement}
Segmentation agreement (IOU) was also good, and further improved through depth-refinement for our proposed system (EDFN: 0.885~$\pm$~0.069, EDFN-D: 0.927~$\pm$~0.029). The graph cut IOU both without and with depth-refinement outperformed our proposed methods owing to user-specified seed points (GC: 0.938~$\pm$~0.036, GC-D: 0.941~$\pm$~0.029). We attribute the improvements of both our proposed system and our applied ground truth with depth-refinement to reduction in the visual-volume discordance due to low-profile yet visually distinct foods on a plate, such as pasta sauce. 

\subsubsection*{Percent Intake Error (2D and 3D)}
Regarding 2D percent intake error (i.e., using segmentation alone), our non-depth-refined and depth-refined proposed systems were outperformed by the depth-refined graph cut implementation (EDFN\subD{2}: -23.4\%~$\pm$~21.2, EDFN-D\subD{2}: -16.5\%~$\pm$~17.6, GC\subD{2}: -6.8\%~$\pm$~13.5, GC-D\subD{2}: -5.3\%~$\pm$~12.7). These negative values, which were improved with depth-refinement, imply a bias towards under-segmenting an image. Refer to the discussion for clinical implications of this bias towards under-segmentation. Regarding the 3D percent intake error, the graph cut implementations outperformed our proposed system (EDFN\subD{3}: -9.1\%~$\pm$~8.8, EDFN-D\subD{3}: -9.0\%~$\pm$~8.9, GC\subD{3}: 0.4\%~$\pm$~1.3, GC-D\subD{3}: 0.4\%~$\pm$~1.3).

\subsubsection*{Volume Estimation Accuracy}
Regarding the volume error (mL) on the regular texture dataset, while initially it appeared depth-refinement worsened performance across methods (EDFN: -14.7 mL~$\pm$~50.0, EDFN-D: -17.2 mL~$\pm$~50.3, GC: 0.0 mL~$\pm$~7.1, GC-D: -1.8 mL~$\pm$~6.9), the plate-level absolute volume error was improved with depth refinement and variance was much smaller (e.g., EDFN-D volume intake error -130.2 mL~$\pm$~154.8; EDFN-D mean absolute error 18.0 mL~$\pm$~50.0). However, intake error from volume was high and both our proposed system and graph cut implementations (EDFN: -129.2 mL~$\pm$~154.3, EDFN-D: -130.2 mL~$\pm$~154.8, GC: 1.8 mL~$\pm$~6.6, GC-D: 0.2 mL~$\pm$~6.5). Corroborated by the negative mean error bias for EDFN and EDFN-D, we empirically attribute this wide variance and high intake error paired with low plate-level absolute volume error to salad. As intake error is calculated with a plate relative to the full portion, variability in a food's appearance could be high leading to high intake error, with accurate absolute volume. Salad, as part of the lunch plates had low-density with widely varying degrees of air pockets between leaves of lettuce at each plating. Depending on how ``fluffy'' the salad was put into position, it could take on differing volumes.  This is discussed in detail later.

\subsection*{Modified Texture Foods}
\subsubsection*{Food Segmentation Accuracy}
Regarding the modified texture foods dataset, food segmentation accuracy was high for our proposed system, however depth-refinement reduced the accuracy based on classical segmentation accuracy calculations (EDFN: 0.922~$\pm$~0.165, EDFN-D: 0.697~$\pm$~0.348). In both cases however, our proposed system outperformed the graph cut analogs (GC: 0.834~$\pm$~0.157, GC-D: 0.656~$\pm$~0.328. Depth-refinement brings the context of volume, whereas the initial ground-truth hand segmentation used for comparison was based solely on visual appearance of food on the plate. Modified texture foods by nature tend to spread out more due to increased fluidity and as such, are at greater risk for the visual-volume discordance.

\subsubsection*{Segmentation Agreement (IOU)}
Compared to the regular texture food dataset, segmentation agreement (IOU) was adequate (over 0.80) and relatively unaffected by depth-refinement. Our proposed system was slightly outperformed by the graph cut implementation on IOU, with the graph cut variance smaller in non-depth-refined segmentation (EDFN: 0.846~$\pm$~0.114, GC: 0.898~$\pm$~0.082). Our proposed method was comparable to graph cut for the depth-refined counterparts (EDFN-D: 0.819~$\pm$~0.166, GC-D: 0.819~$\pm$~0.165).

\subsubsection*{Percent Intake Error (2D and 3D)}

Regarding the 2D percent error of intake, non-depth-refined implementations had unacceptably high percent error of intakes (EDFN\subD{2}: -44.0\%~$\pm$~38.6, GC\subD{2}: -27.4\%~$\pm$~25.5) with unacceptably wide variances still present in depth-refined implementations (EDFN-D\subD{2}: -16.5\%~$\pm$~20.0, GC-D\subD{2}: -14.5\%~$\pm$~18.1). This provides additional evidence to support the need to rectify the visual-volume discordance, especially in modified texture foods where visually salient food remnants are more likely to remain on the plate after consumption. Depth-refinement had a minimal impact on percent intake error for all implementations yielding very low error in estimating volume intake error (EDFN\subD{3}: -0.8\%~$\pm$~5.2, EDFN-D\subD{3}: 1.4\%~$\pm$~5.9, GC\subD{3}: 0.7\%~$\pm$~3.1, GC-D\subD{3}: 2.1\%~$\pm$~4.6). Again, negative intake implications are addressed in the discussion.

\begin{figure*}
    \centering
	\includegraphics[width=.6\textwidth]{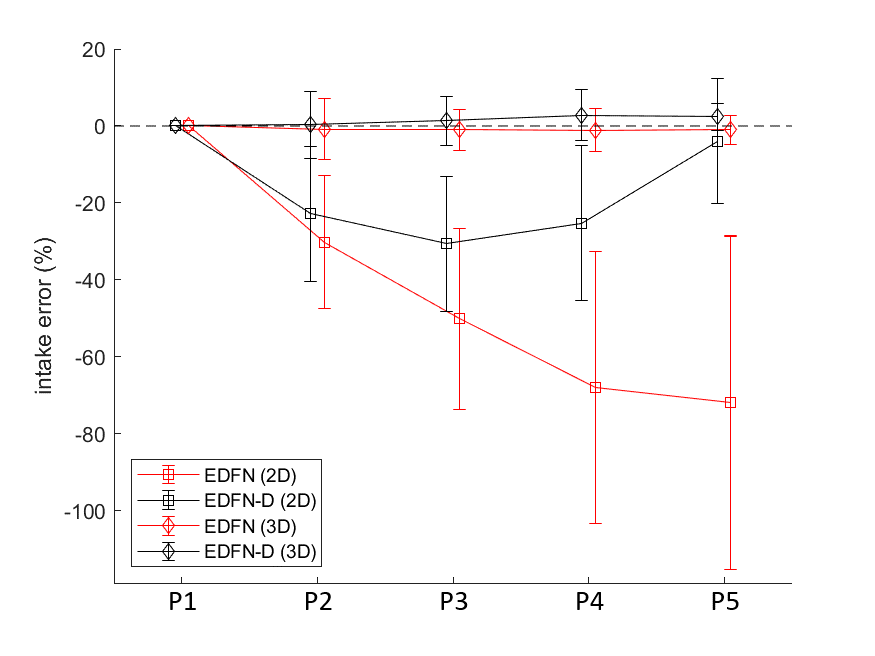}
    \caption{Intake accuracy of our proposed method without (EDFN, red) and with (EDFN-D, black) depth-refinement on the modified texture foods dataset. 2D refers to two-dimensional \% intake error in terms of the proportion of pixels estimated compared to the ground-truth hand segmentation of food areas. 3D refers to three-dimensional \% intake error in terms of the relative volume across the estimated food segment compared to the volume across the ground-truth segmentation of food areas. Here, P1 is the reference plate with P2-P5 representing simulated degrees of increasing intake (i.e., P5 most eaten).}
    \label{fig:intake_error_plot}
\end{figure*}

Figure~\ref{fig:intake_error_plot} depicts these visual-volume discordance errors and illustrates how depth-refinement (black lines) both reduced the variance of the percent intake errors as well as reduced the relative intake error compared to the non-depth-refined counterparts (red) and using P1 as the reference ``full-portion''. Across methods, error tended to increase as the remaining portion size diminished with the exception of EDFN-D (2D) with error peaking at the third portion and receding across portions P4 and P5. We attribute this trend to depth-refinement compensating for higher degrees of visual-volume discordance on plates more likely to have smearing (i.e., plates with less on the plate relative to the initial portion size). This highlights why visual representation context of \textit{where} food resides is inadequate and how additional depth context pertinent to \textit{how much} food is present is required particularly for modified texture foods. 

\subsubsection*{Volume Estimation Accuracy Across Portion Sizes}

\begin{figure*}
    \centering
	\includegraphics[width=.6\textwidth]{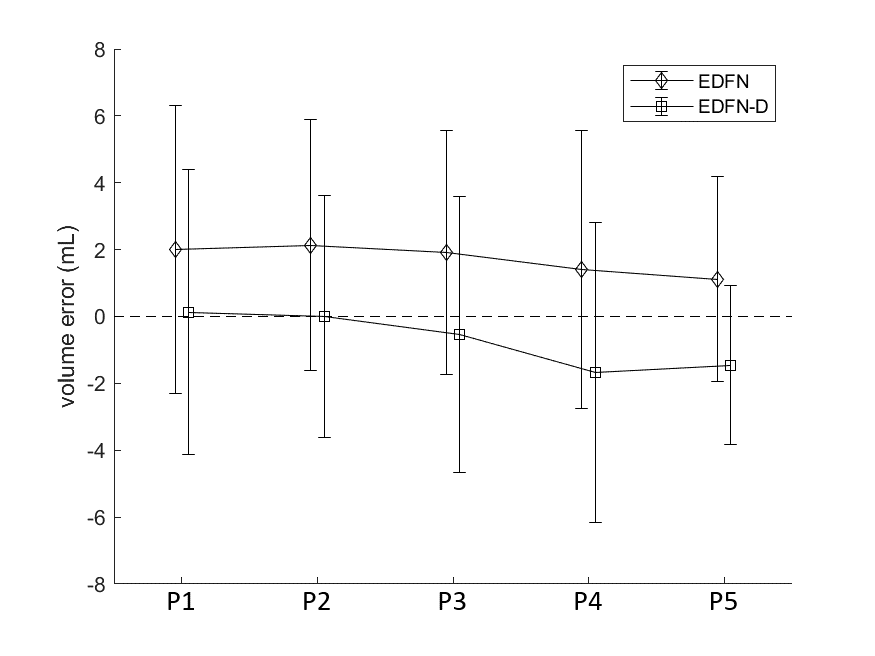}
    \caption{Volume error of the modified texture dataset, in millilitres, of our proposed method without (EDFN) and with (EDFN-D) depth-refinement compared to the volume across the ground-truth hand segmented food area. P1 is the reference plate with P2-P5 representing simulated degrees of increasing intake (i.e., P5 most eaten).}
    \label{fig:vol_error_plot}
\end{figure*}

Typically, volume errors, as well as volume intake errors, were low (less than 4.0 mL) across EDFN and GC implementations as shown in Table~\ref{tab:portion_size_analysis}. To supplement this using our proposed EDFN and EDFN-D, Figure~\ref{fig:vol_error_plot} shows volume accuracy across each of the 5 portions (P1-P5) relative to the volume across the ground-truth hand segmented food regions. While our EDFN-D implementation was more accurate, it had similar precision to EDFN. A similar trend in reduced error in mL across smaller portions was observed for both EDFN and EDFN-D albeit with EDFN-D error shifted downwards. With EDFN-D, the three smallest portions (P3, P4, P5) had negative values indicating the depth-refinement omitted depth values across additional pixels relative to the initial segmentation and yielded a slight under-segmentation of food.

\section*{Discussion}

\label{discussion}

Our proposed fully automatic system imparts a reduced processing burden compared to semi-automatic segmentation. Using graph cuts as our ``applied ground-truth'', task completion time represents an additional point for consideration, and our automated approach provides a key advantage in the food and nutrition tracking context. Empirically, user-defined seed initialization for graph cut implementation incurred approximately 5 seconds of manual annotation time per image. Assuming 192 residents across 6 neighbourhoods (units) in LTC, this implies 48 additional minutes during a meal-service simply to annotate the images. The average time for charting residents' food intake for a day is already at least 270 minutes~\cite{pfisterer2019}, which implies that annotation could impart an 18\% time increase to complete food intake charting. This approach is infeasible and prohibitive within this context. Instead, compared to the graph cuts method which had a tendency to under-segment food, our proposed automatic segmentation method requires minimal additional time commitment from the user enhancing its potential for uptake in the field.

Segmentation errors (under- or over-segmentation) have clinical significance for identification of inadequate intake before low intake progresses to eventual malnutrition. Estimated intake could be incorrect due to either under-estimating food consumption (i.e., less food reported than actually consumed) or over-estimating food consumption (i.e., more food reported than actually consumed). Over-segmenting food areas (i.e., under-estimating intake) implies the predicted food area is over-estimated and translates to reporting that there is more food present (less food intake) than what is true. Conversely, under-segmenting food areas (i.e., over-estimating intake) implies the predicted food area is under-estimated and translates to reporting there is less food present (more food intake) than there is. Over-segmentation (under-estimating food consumption) may be the lesser of two evils for the majority of LTC residents; if food intake is better than what is reported, residents at-risk for malnutrition may be less likely to be missed. Clinically, residents receiving modified texture foods consume significantly fewer calories, have higher cognitive impairment, and require more assistance with activities of daily living than residents on regular texture diets~\cite{keller2018prevalence} making these even higher-risk residents. However, under-segmentation (over-estimating food consumption), increases risk of missing residents with nutrient inadequacies through introducing false negatives and potentially missing residents with poor food intake that could result in malnutrition. While under-reporting may lead to increased referrals for malnutrition screening, most at-risk residents who eat very little would still be identified which might help to identify residents who could benefit from a dietary intervention. 

It appears segmentation is not enough and there is a need for additional context from volume. We observed that under-segmentation was an issue mostly with 2D \% volume intake across implementations and datasets. We evaluated 2D \% volume intake for consistency with existing methods, however, the system's 3D intake estimation is preferred whenever possible. Additionally, returning to Figure~\ref{fig:intake_error_plot}, this plot showcases the need for methods beyond segmentation for food intake tracking. While we observed a decrease in 2D \% intake error with depth-refinement across the remaining plate portions, errors for P2, P3, and P4 were more than 20\%. A system based on segmentation methods alone may be accurate when very little food is consumed or nearly all the food is consumed. Certainly, there is value in improved accuracy for these edge-cases~\cite{parent2012}. However, were we to rely on a system without-depth refinement, this degree of error may be deemed inappropriately high and be a barrier to uptake. Further refinements from depth-refined volume estimations yielded additional improvement in 3D percent error intake and may provide a palatable alternative within an acceptable error margin of less than 10\% (where current practice is up to 62\% error~\cite{castellanos2002}; 50\% for portion size~\cite{bingham1991}) with the added benefit of more fine-grained assessment (continuous measurement versus 25\% incremental bins). While we also observed negative values for volume intake error for our proposed system, EDFN-D, this error was largely due to salad. It appeared as though GC seemed to understand the visual representation of salad better than our proposed method. We suspect this could be improved with additional instances of salad included in the training and validation set. Green vegetables in general were under-represented in UNIMIB2016 and provides an opportunity area for additional comprehensive food intake tracking databases. Whereas the volume intake error for the regular texture foods which included salad was on the order of 130 mL, the modified texture foods circumvented the low density foods problem with volume intake error of 0.8 mL. Arguably, accuracy on these modified texture foods is more clinically relevant for at-risk residents. Additionally, building in some depth redundancy may better approximate a human assessor and improve initial acceptability. 

Perhaps more pertinent to uptake of the system and recurring use of the system is \textit{how} the computer ``sees'' food on a plate and how this compares to the human experience for supporting trust in the automated system. For example, from the human assessor perspective, a plate containing sauce remnants would be ignored and treated as completely consumed food while the computer vision approach would observe each of the pixels containing sauce remnants and mark it as still containing food as in the case of Figure~\ref{fig:vis_comp}. This may lead to distrust in the system because the system makes decisions differently and in opposition to how a human assessor would classify the presence of food. Borrowing from clinicians perceptions of artificial intelligence tools, the alignment between the system output and what would be expected from a human's interpretation is essential for continued use~\cite{tonekaboni2019clinicians}. By incorporating depth-refinement, even though the IOU for modified texture foods decreased, it brings the assessment closer to how a human would interpret a plate with the added benefits of greater precision and objectivity. This area has been largely unexplored because, with one exception, available food datasets only include full plates (see Supplementary Materials for an overview of available food datasets). For measuring food intake, additional application appropriate metrics such as intake error and volume estimation accuracy must be considered since assessing system performance solely from a segmentation accuracy perspective can be misleading.

While our proposed system is not error-free, it is significantly more accurate than current LTC methods. As aforementioned, current LTC home food intake accuracy shows correct estimation of intake occurring as low as 38\% of the time~\cite{castellanos2002} and when portion size is mis-estimated, has error up to 50\%~\cite{bingham1991}. Part of the issue, in practice, may be that the granularity of these estimates is also wide since estimates are recorded as 25\% incremental food intake bins~\cite{med-pass_intake2017,briggs_intake2017}. This may introduce further subjectivity between assessors. Hypothetically, one human assessor may estimate a plate to be 30\% eaten so a value of 25\% would be recorded; another assessor may estimate intake at 45\% reflecting a record of 50\% consumed. With depth-refinement, our proposed EDFN-D removes this subjectivity, operates on a continuous scale, and has a mean 3D \% intake of estimation error of -4.2\% across both datasets and a mean volume intake error of 0.8~mL on the modified texture foods dataset.

To further improve reliability and consistency of the system, negative volume errors and the issue of low-density foods should be considered. For the proposed EDFN-D, the volume errors for plates after the highest intake (P4 and P5) were negative because the depth-refinement omitted volumes across pixels that were initially segmented as food. As part of future work, it would be interesting to explore how necessary initial colour-based segmentation is for establishing \textit{how much} food is present or whether placing greater weight on depth maps could improve accuracy for volume estimation accuracy. Given the issue of low-density foods impacting volume regardless of segmentation method, this must be considered a limitation of over-head food intake systems and these types of foods (e.g., potato chips, salad) may need to be treated differently and separately to other foods. Perhaps repeat imaging of these foods separately, flattening the food before imaging, or applying a general food density score to estimate the range of volume values in these foods could address this limitation.

In summary, we proposed an application-driven design for a novel fully automatic multisensor segmentation system which leverages depth-refinement for improved accuracy. We assessed our system on two representative LTC food intake datasets which included simulated intake plates since current datasets contain only full-plate portions or do not contain pixel-level segmentations. For further advancing the field, additional food intake datasets with pixel-level segmentation are needed. A system such as the one presented here which approximates a human assessor but is objective, faster, more consistent, and can more accurately quantify food intake measurements may provide a valuable step towards automated tracking of food and fluid intake within the LTC sector.

\section*{Methods}

\subsection*{Data Collection}
\label{ssec:procedure}
Data were collected in an industrial research kitchen which conforms to LTC kitchen standards. We constructed an image acquisition system that enabled top-down image capture. We imaged 36 foods representative of LTC where 9 were regular texture foods listed as options on a LTC menu comprising our novel ``regular foods'' dataset. A set of 63 modified texture food samples representing 27 unique foods were prepared by a LTC kitchen (The University Gates, Schlegel Villages) and either minced or pureed comprising our ``modified texture foods'' dataset. During image acquisition, the room temperature varied from 20.6$^{\circ}$C to 22.5$^{\circ}$C. Images were saved to a computer for model training and evaluation. For a summary of the types and representation of foods imaged, see Table~\ref{tab:foods_imaged}.

\begin{table}[]
\caption{Imaged foods represented in each of the regular texture foods dataset and the modified texture foods dataset. Regular texture foods were derived from LTC menus and imaged with three foods per meal (breakfast: oatmeal, toast, scrambled eggs; lunch: tortellini, salad, cookie; dinner: meatloaf, potatoes, corn) at every 25\% incremental amounts of each food item relative to the full portion. Modified texture foods were imaged separately with at five incrementally smaller portions. }
\label{tab:foods_imaged}
\begin{tabular}{lll}
\textbf{Food component}                 & \textbf{Regular texture foods}      & \textbf{Modified texture foods}    \\
\textit{\textbf{Grains}}                &                                     &                                    \\
                                        & Oatmeal                             & Macaroni salad                     \\
                                        & Whole wheat toast                   & Vegetable rotini                   \\
                                        & Cheese tortellini with tomato sauce & Bow tie pasta with carbonara sauce \\
\textit{\textbf{Vegetables and fruits}} &                                     &                                    \\
                                        & Mixed greens salad                  & Sweet and sour cabbage             \\
                                        & Corn                                & Red potato salad                   \\
                                        & Mashed potatoes                     & Seasoned green peas                \\
                                        &                                     & Strawberries \& bananas            \\
                                        &                                     & Stewed  rhubarb \& berries         \\
                                        &                                     & California vegetables              \\
                                        &                                     & Baked polenta/garlic               \\
                                        &                                     & Sauteed spinach \& kale            \\
                                        &                                     & Mango \& pineapple                 \\
                                        &                                     & Asian vegetables                   \\
                                        &                                     & Greek salad                        \\
\textit{\textbf{Proteins}}              &                                     &                                    \\
                                        & Scrambled egg                       & Braised beef liver \& onions       \\
                                        & Meatloaf                            & Teriyaki meatballs                 \\
                                        &                                     & Baked basa                         \\
                                        &                                     & Tuna salad                         \\
                                        &                                     & Hot dog wiener                     \\
                                        &                                     & Braised lamb shanks                \\
                                        &                                     & Salisbury steak \& gravy           \\
\textit{\textbf{Mixed dishes}}          &                                     &                                    \\
                                        & Oatmeal cookie                      & Lemon chicken orzo soup            \\
                                        &                                     & English trifle                     \\
                                        &                                     & Eggplant parmigiana                \\
                                        &                                     & Orange ginger chicken              \\
                                        &                                     & Blueberry coffee crumble cake      \\
                                        &                                     & Barley beef soup                  
\end{tabular}
\end{table}

\subsubsection*{Regular Texture Foods Acquisition}
Three representative meals each consisting of three food items (breakfast: oatmeal, toast, eggs; lunch: pasta, salad, cookie; and dinner: meatloaf, mashed potatoes, corn) were selected from an LTC menu and imaged as part of this data collection series. Each plate was assembled with up to three food items. One full serving of each food item was defined by the nutritional label serving size.
 
Plates were imaged at every permutation of 0\%, 25\%, 50\%, 75\%, 100\% of each food item consumed. Here, 0\% corresponds to the initial, largest mass portion (P1), and 100\% corresponds to no amount of that food component remaining (P5). The largest mass portion, P1 was deemed a ``full'' portion with P2-P5 representing smaller and smaller masses. These 25\% incremental bins were selected based on standard dietary intake record forms used in LTC~\cite{med-pass_intake,briggs_intake}. This yielded  125 unique plates per meal (375 unique plates). Foods were selected to be representative based on a LTC menu.

\subsubsection*{Modified Texture Foods Acquisition}
We imaged 63 food samples (56 unique + 7 duplicates) representing 27 unique food items. Each set of samples for a given unique food contained at least one example of a modified texture (i.e., minced, pureed or both) either imaged fresh, after being held at serving temperature, or both. Holding food at serving temperature is standard practice in LTC serveries as meal items are prepared in advance of meal service. Each food sample was imaged at 5 different portions (P1-P5), with one exception containing 4 portions, by progressively removing some of the sample with a spoon to simulate varying degrees of leftovers for a total of 314 images. The largest mass portion, P1, was deemed a ``full'' portion for intake purposes, with P2-P5 representing smaller and smaller masses. Foods were representative of a typical LTC menu as they were prepared by the LTC kitchen and represented a variety of fruits, vegetables, pastas, soup, and meat dishes.

\begin{sidewaysfigure*}
	\centering
	\includegraphics[width=\textwidth]{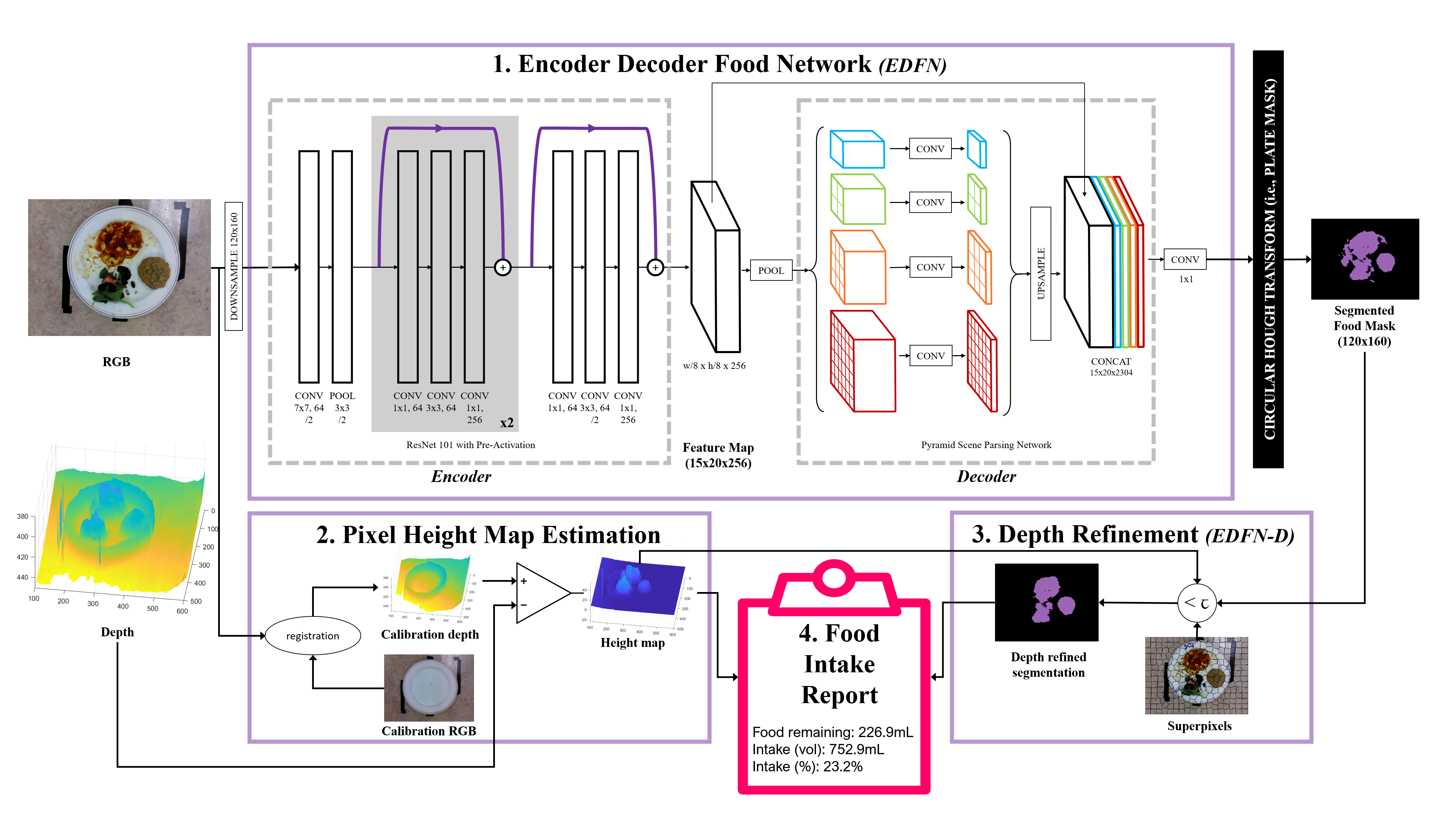}
	\caption{System diagram of the proposed deep food segmentation network comprised of: 1) encoder-decoder food network (EDFN) consisting of a residual encoder microarchitecture~\cite{he2016resnet101} and a pyramid scene parsing~\cite{zhao2017pspnet} decoder microarchitecture which outputs a segmented food mask; 2) pixel height map estimation for assessing food depth; 3) depth refinement which outputs a depth-refined food mask (EDFN-D); and 4) food intake report summarizing the volume of food on the plate, the intake amount (in mL) and the intake percent.}	
\label{fig:network_diagram}
\end{sidewaysfigure*}

\subsection*{Imaging System for Food Volume Estimation System}
The goal was to estimate volumetric food intake from RGB-D images of food on a plate. We developed a deep convolutional neural network (DCNN) for generating food segmentation maps, which was refined using depth heuristics and combined with calibrated pixel-wise food heights to estimate food consumption (in mL). Figure~\ref{fig:network_diagram} shows a visual representation of the system diagram. The following subsections describe the primary subsystems in more detail.

\subsubsection*{Encoder Decoder Food Network (EDFN) Architecture}
\label{sssec:comp_methods}
Inspired by the success of encoder-decoder networks for semantic image segmentation~\cite{badrinarayanan2017segnet}, we designed the macroarchitecture of the proposed food segmentation DCNN as a multi-scale encoder-decoder network architecture tailored for downsampled, pixel-level semantic segmentation of food images. Figure~\ref{fig:network_diagram} shows the network architecture, which consists of a residual encoder microarchitecture, a multi-scale hierarchical decoder microarchitecture, and a final high-resolution, per-pixel classification layer for producing a food segmentation map. The residual encoder microarchitecture is responsible for encoding RGB images into a set of feature maps describing the objects in the image. The encoder feature map outputs are then processed through the decoder microarchitecture which parses the scene at multiple spatial scales. These multi-scale representations were concatenated to the feature map outputs, and a 1x1 convolutional layer was trained to output a two-class per-pixel segmentation map.

For the residual \textit{encoder} microarchitecture, we leveraged a spliced ResNet101 architecture with pre-activation~\cite{he2016resnet101}. The ResNet101 architecture was chosen because of its powerful representational capability for learning discriminative feature representations from complex scenes. We leveraged the notion of transfer learning by beginning with a ResNet101 network architecture designed for classification, trained on the ImageNet dataset of natural scenes~\cite{imagenet_cvpr09}, and splicing off the deeper ResNet101 layers to create the final encoder microarchitecture. More specifically, we splice at the third unit of the first residual block~\cite{zhao2017pspnet}, leading to the proposed residual encoding microarchitecture, which encodes 120$\times$160~RGB images into 256 15$\times$20 feature maps. As such, the image was fed through a 7$\times$7 convolutional layer with 64 kernels and a stride of 2. Then, a 3$\times$3 max pool with stride of 2 was performed to downsample the image. These representations were fed through the first ResNet101 block, consisting of 64 1$\times$1 convolution, 64 3$\times$3 convolution, and 256 1$\times$1 convolution layers three times, with skip connections after every set of 3 layers. The last 3$\times$3 layer was downsampled using a stride of 2. Thus, the encoder microarchitecture outputs 256 feature maps at 1/8 the input image size.

The \textit{decoder} microarchitecture of the proposed food segmentation network was designed to decode the feature maps from the encoder microarchitecture into hierarchical global priors using a region binning scene parsing network architecture design. It is well known that multi-scale context aids pixel segmentation~\cite{liu2019auto} which is particularly relevant within the context of food. As humans observing food, there are two main components: the colour and the texture of the food. Texture also varies across scales (i.e., food has a hierarchical visual nature to it). To account for the multi-scale context of food, we leveraged a pyramid scene parsing network (PSPNet)~\cite{zhao2017pspnet} which was connected to the feature outputs from the encoder microarchitecture. As such, the PSPNet decoder microarchitecture performs analysis across four spatial scales, which adds information representing the underlying feature representation and provides local-to-global context of the plate of food. The feature maps were fed into four parallel max-pool layers, with bin sizes of 1$\times$1, 2$\times$2, 3$\times$3, and 6$\times$6. The upscaled hierarchical global prior outputs were concatenated to the encoder feature maps and two class (food or not food) pixel-level segmentation was performed using a 1$\times$1 convolution layer. A circle Hough transform~\cite{atherton1999} was used to mask the plate from the table, eliminating detection of food outside the plate boundaries (e.g., on tables with complex patterns).

\subsubsection*{Training and Validation Dataset Selection}
\label{sssec:testset_methods}
Evaluating the appropriateness of training and validation dataset selection for application to food intake assessment in LTC requires several considerations: 
\begin{itemize}
    \item \textbf{Colour}: Food comes in many colours. Part of supporting a healthy diet includes the mentality of ``eat a rainbow'' to ensure various micronutrient needs in additional to macronutrient needs are met~\cite{minich2019review}. As such, there should be a wide distribution of colours naturally found in foods captured in the training and validation datasets. 
    \item \textbf{Texture}: Similar to colour, foods also inherently come in a variety of textures. This aspect is particularly salient when considering the LTC population where 47\% of residents receive modified texture diets (e.g., minced, pureed) as part of a strategy to address the high prevalence of swallowing difficulties pervasive in LTC~\cite{vucea2019prevalence}. 
    \item \textbf{Portion}: Given the application to food intake assessment, the ideal training dataset would have representative intake images that include both before and after images of meals (i.e., not just full portions of served food). The rationale is by having more representative examples of what foods can look like partially, or fully disassembled in the case of food mixing, in the training dataset, the network will be more likely to learn representative examples of foods apart from the original served context.
    \item \textbf{Orientation}: The occlusion conundrum where one food occludes another is very difficult to circumvent especially in the LTC environment when taking multiple images from many perspectives is infeasible due to time constraints. For the purpose of acquisition ``in the wild'' within LTC, images taken from the above configuration is preferred to facilitate volume estimation and down-stream nutritional intake estimation. The issue of occlusion (e.g., seeing only the bun as part of an assembled hamburger) remains an issue, however is reasonable when accepting the assumption that complex foods (e.g., foods with multiple components like a hamburger) are eaten in similar proportions. While this is an undoubtedly fallible assumption, errors in down-stream nutritional estimation are constrained to a specific food as opposed to influencing the entire plate. As such, the ideal training and validation database would be acquired in the top-view configuration. 
    \item \textbf{Label level}: To facilitate volume estimation and down-stream nutrient estimation, image segmentation must be conducted pixel-wise as opposed to using bounding boxes. As such, the ideal training and validation database would be labelled at the pixel-level. 
    \item \textbf{Style}: Multiple food items on a plate is common in LTC as the ``family style'' approach to eating has been shown to enhance food intake~\cite{vucea2014interventions}. While multiple plates may also be used in LTC (e.g., soup, side salad, dessert), the training and validation dataset would ideally show representation consistent with ``family style'' as opposed to solely ``single item'' plates. 
    \item \textbf{Accessibility}: The ideal dataset needs to be readily available (i.e., non-proprietary). 
\end{itemize}
Based on the above considerations, the most imperative are orientation and label level and the most suitable food database available for training and validation for LTC (i.e., top-view with pixel-wise labelling) is therefore UNIMIB2016~\cite{cioccaJBHI}. Refer to Supplementary Materials for a compilation of popular food databases with a summary of rationale for selecting UNIMIB2016.

\subsubsection*{Training}
\label{sssec:training}
We trained the proposed encoder-decoder food network (EDFN) on the UNIMIB2016 food dataset (1027 tray images, 73 categories), which contains per-pixel ground-truth segmentation~\cite{cioccaJBHI}. The encoder weights were frozen to conserve deep computational feature extraction from large robust datasets, and only the decoder weights were optimized. The UNIMIB2016 dataset was chosen due to its food variety, overhead view, and pixel-level segmentation annotation. Additionally, since our method was driven by LTC application requirements with data collection in a specific manner, we needed a dataset that was similarly acquired (e.g., pixel-wise annotation, not bounding boxes) so training/fine-tuning could be accomplished without bias by our novel LTC test datasets. While our LTC test datasets comprise a representative sample albeit with relatively few food groupings, training on the UNIMIB2016 dataset with 76 food categories enhanced generalizability. Downsampling was conducted to align the spatial feature sizes of the encoder and decoder microarchitectures with the UNIMIB2016 dataset~\cite{cioccaJBHI}.  The UNIMIB2016 data were resized to match our image height/width which were at the same aspect ratio (4:3). This resizing to 120x160 images provided two key advantages: (1) computation reduction, (2) better scaled kernels for the image size. We empirically observed that there was not enough global context at the original resolution, resulting in the middle of foods getting misclassified. By downsampling our image, the network was able to identify primary low-level features instead of getting stuck in the texture of the food and could be successfully decoded by the pyramid scene parsing decoder microarchitecture. The UNIMIB2016 data were randomly split into training and validation subsets (80\%/20\%). Since all UNIMIB2016 plates were placed on the same tan colored tray, we found that the machine learning model inappropriately learned that the tray colour was always indicative of non-food. Thus, we performed data augmentation on the UNIMIB2016 data by randomly rotating the hue channel of each image's background (non-food) pixels and adding it to the dataset, thus effectively doubling the training and validation datasets. The network was trained using batch size 32 using RMSProp optimizer with softmax cross-entropy loss, a learning rate of 0.0001 and a decay of 0.995. The network was trained over 200~epochs, and the best model according to the validation loss was kept.

\subsubsection*{Segmentation Depth-Refinement}
\label{sssec:depth_refinement}
The generated food segmentation map from EDFN is based solely on visual information, and is thus privy to visual-volume discordance. We therefore developed a heuristic for excluding labeled food areas that are irrelevant to food consumption (e.g., pasta sauce remnants). To do this, co-aligned depth maps were acquired synchronously with the RGB images for the plate under analysis as well as an empty calibration plate.

Ten depth maps were averaged for each acquisition to account for measurement noise. The calibration depth map $d_C$ was registered to the plate depth map $d_P$ to account for any changes in camera-plate orientation between calibration and plate acquisitions. To accomplish this, the plate edges were identified from the RGB plate images using a Canny edge detector with a Gaussian filter of $\sigma$=3 and hysteresis threshold values of 10 and 50. A circle Hough transform, using the range of expected plate radii, was performed on these edge maps to determine the plate center and radius. Denoting the plate centers of the calibration and food plate as $(x_C, y_C)$ and $(x_P, y_P)$, a translation transformation $T=(x_P-x_C, y_P-y_C)$ was applied to $d_C$, thus aligning the two depth maps. Pixel-wise food height was then computed:
\begin{equation}
    h_i = d_{C,i}-d_{P,i}
\end{equation}
where $h_i$ is the food height relative to the plate for pixel $i$ in mm. The transformation $T$ accounts for planar translation between acquisitions, but changes in tilt may also affect the measurement. Thus, a full-field correction was performed by constraining the left and right limits of $h_i$ (table regions) to 0~mm. Specifically, for each row, we subtracted the weighted average of the left and right table pixel heights (which should theoretically be 0~mm), weighted to the pixel's distance from the left and right boundaries. A $5\times5$ median filter was applied to the food height map to correct spurious measurements at plate boundaries.

Food height was used to refine the segmentation mask (``depth-refinement'') based on \textit{a priori} knowledge that visual-volume discordance is observed when very shallow and inconsequential foods are visually apparent, but are irrelevant to volumetric analysis. Specifically, the image was decomposed into 250 perceptually meaningful superpixels using simple linear iterative clustering~\cite{achanta2012} (compactness=20, $\sigma$=2). For each superpixel $\mathcal{S}_i$, the constituent pixels were removed from the food map using statistical thresholding on the pixel height distribution:
\begin{equation}
    Q_{h_{\mathcal{S}_i}}(p) < \tau
\end{equation}
where $Q_{h_{\mathcal{S}_i}}$ is the quantile function of the distribution of pixel heights in $\mathcal{S}_i$. We set $p=0.75$ and $\tau=2$~mm based on measurement error along a flat table.

\subsubsection*{Food Volume Calculation}
Food height was determined in mm units, but to calculate food volume, pixel spacing needed to be calibrated to mm (a pixel-to-mm conversion). Using the known diameter of the plate $d$ (259~mm), we used the detected plate radius $\hat{r}$ from the circle Hough transform to compute the conversion:
\begin{equation}
    \Delta x = \frac{d}{2\hat{r}}
\end{equation}
Food volume in mL could then be computed by summing the per-pixel differential volumes within the food mask:
\begin{equation}
    V = \sum_{i \in F} (\Delta x)^2 h_i
\end{equation}
where $F$ is the set of segmented food pixels. Volumetric food intake was computed by subtracting the plate volume from the full portion volume. Similarly, percent intake was calculated relative to the full portion volume.

\subsubsection*{Testing}
\label{sssec:testing}
We tested the network on our two custom LTC datasets consisting of 689 (375+314) images representing 36 (9+27) different foods (as outlined in Table~\ref{tab:foods_imaged}). Original images were downsampled from 480$\times$640 to 120$\times$160 to decrease the number of network parameters and improve computation time. The images were hand segmented to define ground truth segmentation masks of the food on the plates.

\begin{figure}
	\centering
	\includegraphics[width=\linewidth]{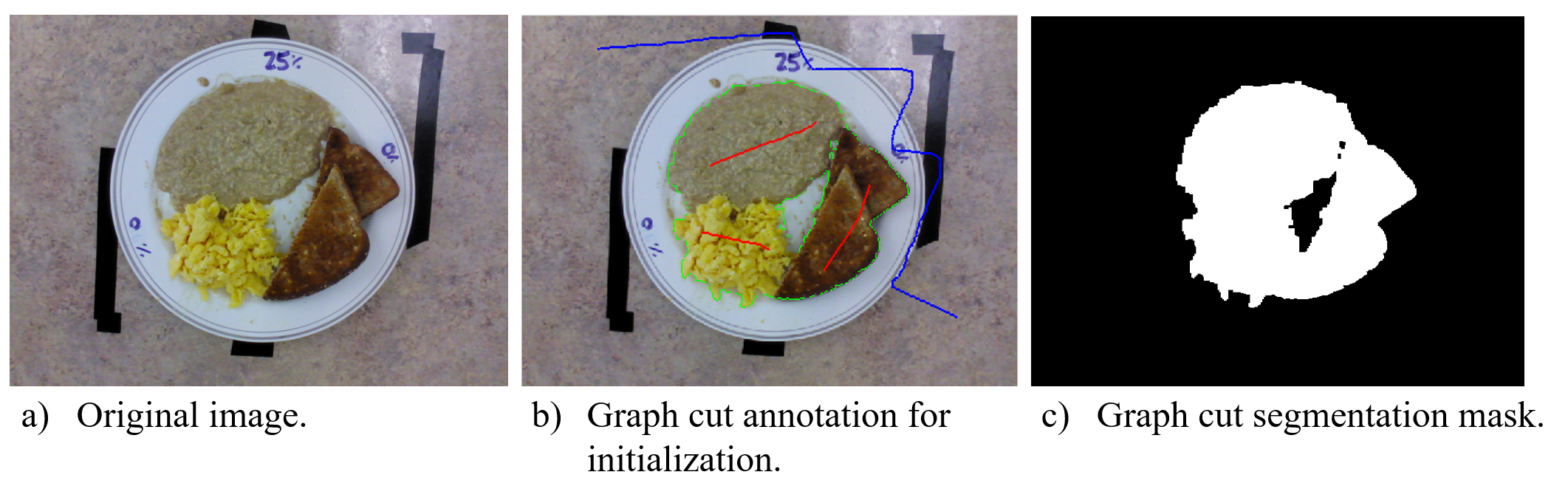}
		\caption{Sample graph cuts annotation with one line per food item (red) and one background line (blue) and resulting segmentation mask.}	
    \label{fig:GC_annotation}
\end{figure}

We compared our results to those generated by semi-automatic graph cut segmentation. Since user input is required for initialization, for consistency in the regular texture dataset, one line was used to denote each food item present on the plate and one squiggled background line was indicated around the top and right side of the image as shown in Figure~\ref{fig:GC_annotation}. The modified texture dataset required additional user-defined seeding. The circle Hough transform plate masking used in our proposed system was used here too. The output from this method is a plate-level food segmentation mask.

\subsection*{Data Analysis}
\label{sssec:analysis}
To compare quantitative performance between methods, we use the common performance measures of global accuracy (Equation~\ref{eq:1}) to describe the percentage of correctly classified pixels, food segmentation accuracy (Equation~\ref{eq:2} to describe the percentage of correctly classified food pixels, as well as the intersection over union (IOU) (Equation~\ref{eq:3}) both within a meal (i.e., breakfast, lunch, dinner, modified texture single-imaged foods) and across meals. For this application, the IOU provides a more representative metric for how our segmentation system is performing as it captures accuracy within the context of the true bounded food areas since false positive predictions are penalized. The theoretical maximum value of IOU is 1.0 when the intersection maps perfectly over the union without deviation. We define the  metrics described above as follows:
\begin{align} \label{eq:1}
    \text{Global Accuracy} &= \frac{TP + TN}{TP + TN + FP + FN} \\ \label{eq:2}
    \text{Food Segmentation Accuracy} &= \frac{TP}{TP+FN} \\  \label{eq:3}
    IOU &= \frac{target \cap prediction}{target \cup prediction} 
\end{align}

Volume estimation accuracy was assessed by computing mean absolute intake error (mL) as volume calculated using our proposed method or the ``applied ground-truth'' method with or without depth-refinement relative to the volume across the ground-truth hand segmented areas. Error (mL) was calculated similarly but preserves the direction of error. Volume intake error (mL) is the difference between the current portion relative to the full portion. Intake error was calculated for both segmentation (2D) and volume (3D) data relative to the full portion. All values are reported as mean~$\pm$~SD.

\bibliography{bib}

\section*{Acknowledgements}
The authors would like to acknowledge Yitian Wang for her contribution to manual food segmentation. This work was funded by the National Science and Engineering Research Council (NSERC), and the Canada Research Chairs (CRC) program.

\section*{Author contributions statement}

K.J.P and R.A contributed equally to this work. K.J.P conceptualized the system; R.A and A.W provided additional contributions to system design. K.J.P was the main contributor to experimental design, and contributed to algorithmic design. K.J.P was the main contributor for data acquisition protocols and data collection with contributions from R.A and A.G.C. R.A was the main contributor to algorithmic design with contributions from B.S, A.M, and A.W. K.J.P was the main contributor to data analyses; K.J.P and R.A. conducted data analyses. H.H.K provided the clinical nutrition perspective and direction, and facilitated and oversaw data collection in the test kitchen. K.J.P was the main contributor to writing the manuscript. All authors reviewed the manuscript.

\end{document}